\documentclass[10pt,twocolumn,letterpaper]{article}

\usepackage{cvpr}      

\newif\ifonlyappendix

\usepackage[table]{xcolor}
\definecolor{mypink}{HTML}{d896b3}
\definecolor{myorange}{HTML}{ff8d2b}
\definecolor{myblue}{HTML}{1f59a1}
\definecolor{mygreen}{HTML}{3b93bb}
\definecolor{mygrey}{HTML}{aaaa9e}
\definecolor{myyellow}{HTML}{f0c369}
\definecolor{mypurple}{HTML}{7f439c}
\definecolor{myspike}{HTML}{1e9695}

\usepackage[labelfont=bf]{caption}
\usepackage{booktabs}
\usepackage{multirow}
\usepackage{bm}
\usepackage[most]{tcolorbox}
\usepackage{colortbl}

\newcommand{\data}{VCG-32K\xspace}
\newcommand{\model}{CauSight\xspace}

\makeatletter
\renewcommand\paragraph{\@startsection{paragraph}{4}{0pt}%
  {0.5ex plus 0.2ex minus 0.2ex}
  {-1em}
  {\normalfont\normalsize\bfseries}}
\makeatother

\usepackage{changepage}
\usepackage{relsize, graphicx}

\definecolor{cvprblue}{rgb}{0.21,0.49,0.74}
\usepackage[pagebackref,breaklinks,colorlinks,allcolors=cvprblue]{hyperref}

\usepackage{xcolor}

\def\paperID{1705} 
\def\confName{CVPR}
\def\confYear{2026}

\title{\model: Learning to Supersense for Visual Causal Discovery}

\author{
Yize Zhang$^{1,2,3}$\thanks{Equal contribution.} \quad
Meiqi Chen$^{4}$\footnotemark[1] \quad
Sirui Chen$^{1,5}$\footnotemark[1] \quad
Bo Peng$^{1,2,3}$\footnotemark[1] \\
Yanxi Zhang$^{1,4}$ \quad
Tianyu Li$^{2}$ \quad
Chaochao Lu$^{1}$\thanks{Corresponding author.} \\
{$^{1}$Shanghai AI Laboratory}\quad
{$^{2}$Shanghai Innovation Institute}\\
{$^{3}$Shanghai Jiao Tong University}\quad
{$^{4}$Peking University}\quad
{$^{5}$Tongji University}\\
{\tt\{zhangyize, luchaochao\}@pjlab.org.cn}
}

\begin{document}
\maketitle

\ifonlyappendix
\else
  \begin{abstract}
Causal thinking enables humans to understand not just what is seen, but why it happens. 
To replicate this capability in modern AI systems, we introduce the task of \textbf{visual causal discovery}. It requires models to infer cause-and-effect relations among visual entities across diverse scenarios instead of merely perceiving their presence.
To this end, we first construct the Visual Causal Graph dataset (\data), a large-scale collection of over 32,000 images annotated with entity-level causal graphs, and further develop \model, a novel vision-language model to perform visual causal discovery through causally aware reasoning. 
Our training recipe integrates three components: (1) training data curation from \data, (2) Tree-of-Causal-Thought (ToCT) for synthesizing reasoning trajectories, and (3) reinforcement learning with a designed causal reward to refine the reasoning policy. 
Experiments show that \model outperforms GPT-4.1 on visual causal discovery, achieving over a threefold performance boost (21\% absolute gain).
Our code, model, and dataset are fully open-sourced at project page:  \href{https://github.com/OpenCausaLab/CauSight}{https://github.com/OpenCausaLab/CauSight}.

\end{abstract}

  \section{Introduction}

\begin{figure}[ht]
    \centering
    \includegraphics[width=0.85\linewidth]{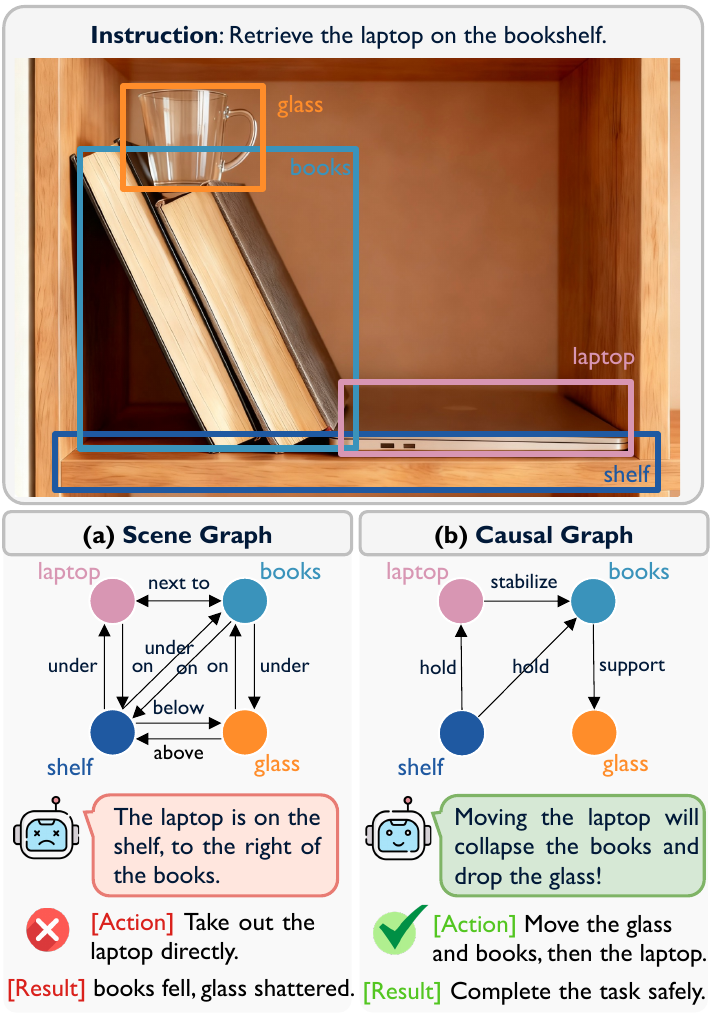}
    \caption{\textbf{A comparison between VLMs that understand (a) scene graph, which only specifies spatial relations between entities; (b) causal graph, which captures causal mechanisms between entities.} Genuine reasoning and safe deployment require VLM to discover causal relations between entities.
    }
    \vspace{-4mm}
    \label{fig:1}
\end{figure}

Causal thinking is regarded as a hallmark of human intelligence~\citep{Penn2007CausalCI, Udbhav2017SapiensAB, Schulz2007SeriousFP, Peters2017ElementsOC}. 
Emulating this capacity on modern vision-language models (VLMs)~\citep{Bordes2024AnIT, Li2025ASO, Awadalla2023OpenFlamingoAO, Zhang2023VisionLanguageMF, Zhu2023MiniGPT4EV, Dai2023InstructBLIPTG} is core to achieving not just recognition, but genuine reasoning about our visual world~\citep{Ke2021SystematicEO, Krishna2016VisualGC, Zellers2018FromRT, Nair2019CausalIF, Bengio2019AMO, Kurutach2018LearningPR, Chalupka2014VisualCF}.
For instance, as \autoref{fig:1} shows, the underlying causal graph provides a clear mechanistic narrative for an embodied agent tasked with retrieving a \texttt{laptop}: 
the initial action must be to stabilize the \texttt{glass}, since directly removing the \texttt{laptop} would trigger a causal chain reaction, collapsing the \texttt{books} and consequently dropping the \texttt{glass}.
These causal graphs~\citep{Abdulaal2024CausalMA, Jiralerspong2024EfficientCG} could support logical, safe, and precise downstream decision-making with robust generalization, even under changing conditions~\citep{yu2025adam}.

To discover such causal graphs across diverse real-world visual scenarios, primary steps involve identifying causal entities and inferring their pairwise causal relationships.
We follow the definition of causality in the vision-language context based on the concept of ``causal dispositions''~\citep{Mumford2011GettingCF, LopezPaz2016DiscoveringCS, Chen2024CELLOCE}, to describe a causal relationship as existing when one entity inherently possesses the ability to influence the state of another.
This can be further illustrated through counterfactual reasoning—if the \texttt{cause} were absent, the \texttt{effect} would not sustain its current state~\citep{Frappier2018TheBO, Pearl2013StructuralCA}.
We thus position \emph{visual causal discovery} as the problem of grounding visual concepts in language through explicit reasoning about their causal relationships.
However, this very task remains a significant and ongoing challenge in the field.

On one side, previous efforts have rarely focused on causal modeling of scenes in the vision-language context.
The landscape of scene graph research~\citep{Chang2021ACS, Johnson2015ImageRU, Yang2022PanopticSG} was oriented toward object detection and predicate recognition, remaining agnostic to the underlying causality.
As illustrated in \autoref{fig:1}, conventional scene graphs merely depict static adjacency or capture associative structures among entities, without causally modeling how changes in one affect another.
This inherent limitation severely restricts their value for downstream reasoning and intervention tasks.
In contrast, causal graphs explicitly reveal the chain of cause-and-effect relationships among entities, offering a dynamic, interactive, and unambiguous representation of visual scenes.

On the other side, even the most advanced VLMs currently struggle with the fundamental task of visual causal discovery~\citep{Chen2024CELLOCE}.
It requires the model not only to identify entities and their attributes, but also to infer causal relationships via counterfactual reasoning, while eliminating confounders.
Based on our extensive testing in \autoref{sec:exp}, the average recall rate of causal graphs for open-source models like Qwen2.5-VL series~\citep{Bai2025Qwen25VLTR} is only \textbf{10.0\%}, even proprietary models such as OpenAI o3~\citep{OpenAIOA} and GPT-4.1~\citep{Achiam2023GPT4TR} achieve merely \textbf{7.3\%} and \textbf{10.2\%}, respectively.
Such performance hinders the practical use of causal graphs in applications such as robotic manipulation and autonomous driving that require safe and reliable decision-making~\citep{wang2023openlanev2, li2023toponet}.

To address the aforementioned bottlenecks, we introduce the Visual Causal Graph dataset (\data), the first large-scale collection of over 32,000 images annotated with entity-level causal graphs.
VCG-32K is built upon 11,428 samples from MS-COCO~\citep{Lin2014MicrosoftCC} and 20,828 from Objects365~\citep{Shao2019Objects365AL}, both widely serving as major sources for visual understanding.
Guided by the definition of visual causality described above, we employ a systematic annotation process to ensure consistency and reliability, enabling rigorous benchmarking on \emph{visual causal discovery}.

We further develop \model, a novel VLM capable of generating underlying causal graphs across diverse visual scenarios through causally aware reasoning.
Our training recipe consists of three components: (1) training data curation from the proposed \data dataset; (2) Tree-of-Causal-Thought (ToCT), an automated approach for synthesizing visually grounded reasoning trajectories; (3) reinforcement learning (RL)~\citep{Chen2021DecisionTR, sheng2024hybridflow} with our designed causal reward to optimize the model's policy for visual causal discovery. 
Specifically, for training data curation, we use the MS-COCO portion of the \data dataset as the training corpus, enabling cross-dataset evaluation on the Objects365 subset. 
As these data offer only final labels but no process supervision, we propose ToCT to synthesize high-quality trajectories: a teacher model\footnote{We use Qwen2.5-VL-72B for synthesizing trajectories for its strong reasoning performance and low cost compared with proprietary models.} repeatedly executes actions of \emph{region selection}, \emph{entity recognition}, and \emph{causality orientation} to perform multi-step reasoning.
Monte Carlo Tree Search (MCTS)~\citep{Browne2012ASO} is employed to ensure the quality of these trajectories by maintaining multiple branches and propagating value estimates~\citep{Zhang2025ARiseTK, Yao2023TreeOT, Zhang2024ReSTMCTSLS}.
The synthesized trajectories are then filtered and serve as demonstrations for the policy model\footnote{We use the term \emph{policy model} to refer to the model being trained; in this work, we adopt Qwen2.5-VL-7B as our base model for its favorable balance between capability and training cost.} to develop an initial reasoning policy through supervised fine-tuning (SFT)~\citep{Ouyang2022TrainingLM, zheng2024llamafactory}. 
Lastly, we refine the policy model using the Group Relative Policy Optimization (GRPO) algorithm~\citep{Shao2024DeepSeekMathPT} guided by a carefully designed, graph-based causal reward, which encourages accurate visual causal discovery and faithful structural generation.
Experimental results demonstrate that our model, \model, achieves superior performance, surpassing its base model and the leading proprietary GPT-4.1 by absolute gains of \textbf{27.4\%} (\emph{$\approx$ 8.2$\times$} improvement) and \textbf{21.0\%} (\emph{$\approx$ 3.1$\times$} improvement), respectively.
Additionally, \model exhibits strong cross-dataset generalization.

The main contributions of our work are three-fold:
\begin{itemize}
    \item[--] We formulate the task of \emph{visual causal discovery} and introduce the \data dataset, a large-scale collection of over 32,000 images annotated with entity-level causal graphs to facilitate research on this problem.

    \item[--] We introduce the Tree-of-Causal-Thought (ToCT) approach for synthesizing high-quality reasoning trajectories, combined with a reinforcement learning framework optimized by a graph-based causal reward to enhance visual causal discovery, resulting in a novel VLM \model.
    
    \item[--] \model shows superior performance on visual causal discovery, with markedly improvements over baseline models and strong cross-dataset generalization. Further analysis provides valuable insights for future research.
    
\end{itemize}

\section{Task Formulation: \emph{Visual Causal Discovery}}
\label{sec:task_define}

Visual causal discovery aims to construct a visually grounded causal graph $\mathcal{G} = (\mathcal{V}, \mathcal{E})$ from a single image, which consists of two key components: basic entities $\mathcal{V}$ and their causal relationships $\mathcal{E}$.

\noindent\textbf{Entities.} 
Each graph $\mathcal{G}$ contains $N$ entities $\mathcal{V} = \{\mathbf{v}_i\}_{i=1}^{N}$, with $\mathbf{v}_i = (c, b)$ where $c$ is the category label and $b\in\mathbb{R}^4$ is the bounding box.

\noindent\textbf{Causal Relationships.} 
Edges $\mathcal{E}= \{\mathbf{e}_{ij}\}_{i\neq j}$ are directed and predicate-labeled: $\mathbf{e}_{ij} = (\mathbf{v}_i, \kappa, \mathbf{v}_j)$, $\kappa \in \mathcal{K}$ (e.g., ``support''). Following causal dispositions \citep{Mumford2011GettingCF, LopezPaz2016DiscoveringCS, Chen2024CELLOCE}, $i\to j$ exists if an intervention on $\mathbf{v}_i$ changes $\mathbf{v}_j$:
\begin{align}
    \mathbf{e}_{ij}\in\mathcal{E} \Longleftrightarrow 
    p(\mathbf{v}_j \mid do(\mathbf{v}_i = 0))\neq p(\mathbf{v}_j).
\end{align}
Here we let $\mathbf{v}_i\in\{0,1\}$ denote whether an entity exists, where $\mathbf{v}_i=1$ means present and $\mathbf{v}_i=0$ means absent. In this formulation, $p(\mathbf{v})$ denotes an entity state (e.g., presence, sit). Interventions $do(\cdot)$ here represent \emph{removal} of the entity (i.e., setting $\mathbf{v}_i=0$). For example, as shown in \autoref{fig:1}, if we \texttt{remove} the \texttt{laptop}, the states of \texttt{books} and \texttt{glass} will be changed.

\noindent\textbf{Evaluation.} 
Predicted graphs are evaluated structurally, ignoring entity or predicate labels. Entities are matched to ground-truth via the Hungarian algorithm~\citep{Kuhn1955TheHM} with GIoU~\citep{Rezatofighi2019GeneralizedIO}. An edge $(\mathbf{v}_i, \mathbf{v}_j)$ is correct if it exists in the ground-truth with the same causal direction $i\to j$.

  \section{The \data Dataset} 
Existing visual understanding datasets, such as those focused on scene graphs~\citep{Chang2021ACS, Johnson2015ImageRU, Yang2022PanopticSG}, remaining agnostic to the underlying causality. They provide spatial relations (e.g., ``on") but miss the inner causal mechanism (e.g.,  ``support", ``lift"). Moreover, these datasets suffer from bounding box annotation errors. To address the gaps, we construct the Visual Causal Graph dataset (\data) with three key features: (1) refined bounding boxes for accurate localization of entities $\{\mathbf{v}_i\}_{i=1}^{N}$, (2) explicitly annotated causal relationships $\mathbf{v}_i\to \mathbf{v}_j$, and (3) causal mechanism types $\kappa$ explaining ``why" the causal relationship exists. These rich annotations guide models toward causally aware reasoning beyond perception-level visual understanding.

\data is construced with images from the MS-COCO~\citep{Lin2014MicrosoftCC} and Objects365~\citep{Shao2019Objects365AL} datasets. We employ 50 trained annotators for dataset construction.
\autoref{fig:dataset_pipeline} illustrates the two-stage annotation pipeline: bounding box refinement and causal relationship labeling.  
In the first stage, annotators refine the existing bounding boxes. They delete or modify boxes with incorrect boundaries, label errors (e.g., overly broad, abstract concepts, or incorrect entity names), and entities without direct contact with any other entities. Annotators are also allowed to add new entities that have causal relationships with others.
In the second stage, annotators proceed to identify and categorize causal relationships $\mathbf{e}_{ij} = (\mathbf{v}_i, \kappa, \mathbf{v}_j)$. A causal relationship $i\to j$ exists when three conditions are met: (1) $\mathbf{v}_i$ is in direct contact with $\mathbf{v}_j$, (2) the presence of $\mathbf{v}_i$ maintains the current state of $\mathbf{v}_j$, (3) removing $\mathbf{v}_i$ would cause $\mathbf{v}_j$ to lose its current state. 
We conduct quality checks throughout both stages to ensure the average annotation accuracy exceeds 95\%: 10 reviewers randomly sample items, evaluate annotation accuracy, and correct any errors.
In total, the resulting dataset comprises 32,256 images, 299,262 entities, 2,287 entity categories, and 185,321 causal relationships (5.75 relationships per image in average). For more details, see the Appendix.

\begin{figure}[t]
    \centering
    \includegraphics[width=0.8\linewidth]{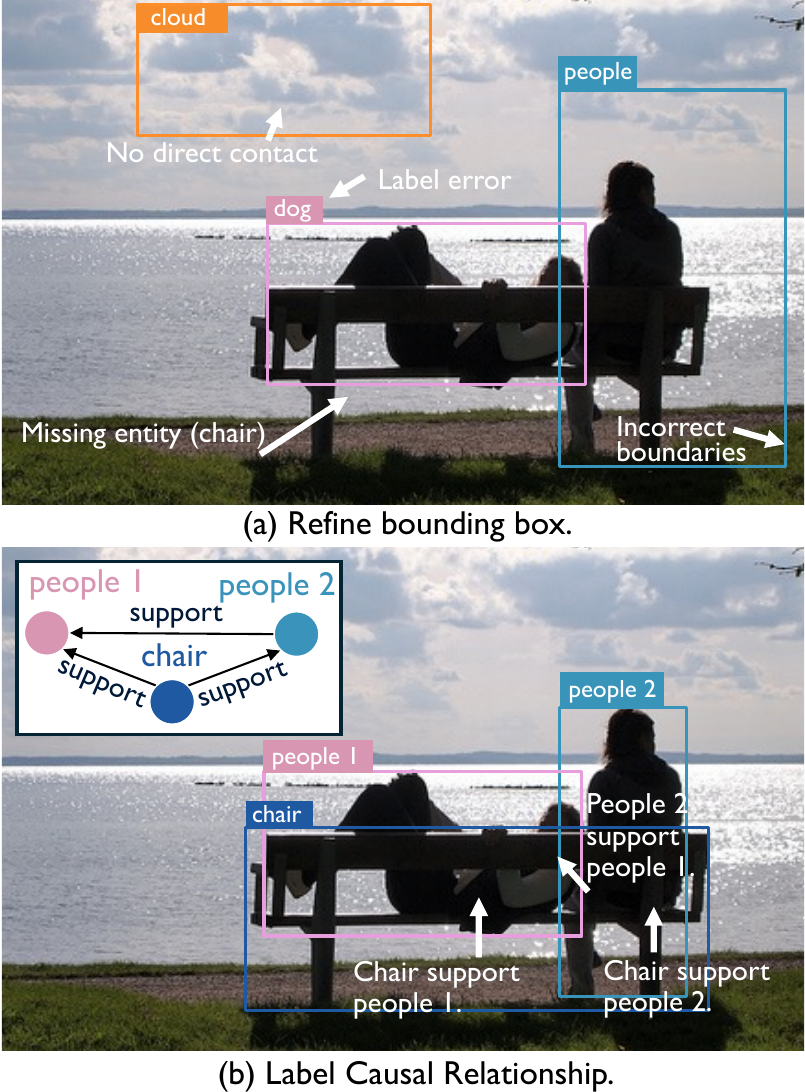}
    \caption{\textbf{The two-stage annotation pipeline of \data}: bounding box refinement and causal relationship labeling.}
    \vspace{-4mm}
    \label{fig:dataset_pipeline}
\end{figure}
  \begin{figure*}[!t]
    \centering
    \includegraphics[width=0.81\linewidth]{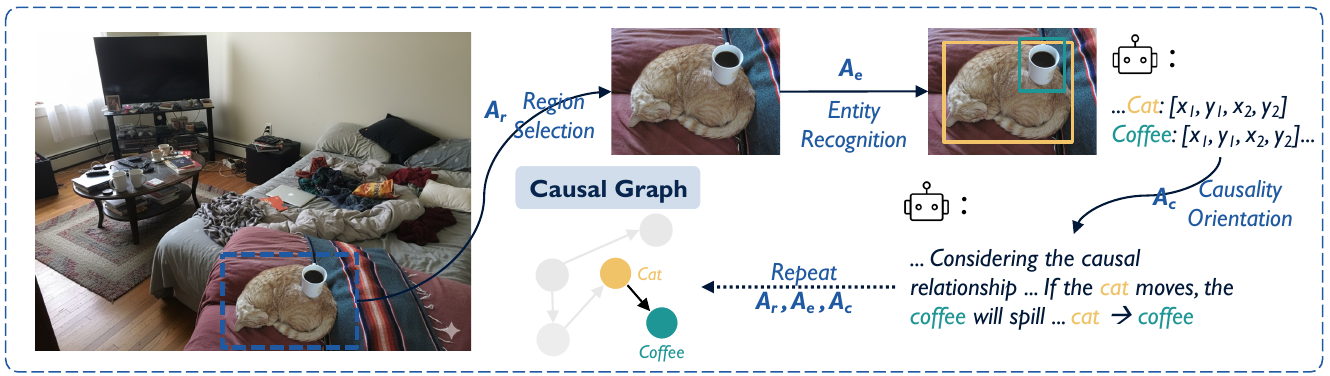}
    \caption{\textbf{Illustration of a single synthesized reasoning trajectory.} The teacher model can repeatedly execute three key actions to extend the reasoning trajectory.}
    \label{fig:traj}
    
    \vspace{2mm} 

    \includegraphics[width=0.81\linewidth]{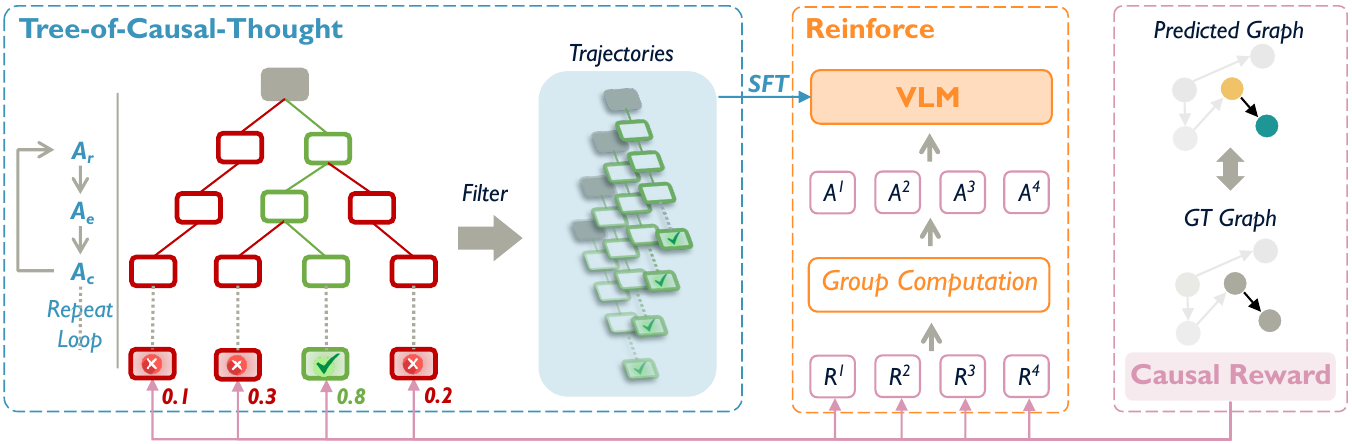}
    \caption{\textbf{Overview of the training pipeline.} Tree-of-Causal-Thought expands a single reasoning path into a tree structure to select high-quality trajectories, which serve as demonstrations to initialize \model's reasoning policy. The subsequent RL further optimizes this policy for visual causal discovery with our designed causal reward.}
    \label{fig:pipeline}
    
    \vspace{-4mm}
\end{figure*}

\section{Method}
In this section, we present the recipe for developing \model. We start by introducing a tree-search-based trajectory synthesis approach for initializing the reasoning policy in \autoref{subsec:toct}, followed by a reinforcement learning stage for further optimization in \autoref{subsec:rl}. \autoref{fig:traj} illustrates part of a single synthesized reasoning trajectory, and \autoref{fig:pipeline} provides an overview of the training pipeline.

\subsection{Tree-of-Causal-Thought}
\label{subsec:toct}

Visual causal discovery seeks to ground visual concepts in language, thereby requiring VLMs to internalize a causally aware reasoning process. However, as \data offers final answers but no process supervision, fine-grained reasoning trajectory cannot be reliably acquired; collecting expert rationales is prohibitively costly and inefficient. 
Consequently, we propose \textbf{Tree-of-Causal-Thought (ToCT)}, an automated approach for synthesizing high-quality causally aware reasoning trajectories. We first present the key actions, then the MCTS procedure, and finally how the resulting small ToCT dataset supervises model training.

\paragraph{Key Actions.}
Constructing ToCT involves three key actions: \emph{region selection} $A_r$, \emph{entity recognition} $A_e$, and \emph{causality orientation} $A_c$. (1) \emph{Region selection} zooms into specific sub-regions, facilitating a more comprehensive search for causal pairs and helping to exclude confounding factors. (2) \emph{Entity recognition} grounds entities in the selected region. (3) \emph{Causality orientation} infers the existence and direction of causal relationships between entities.
The actions are repeatedly executed in a fixed loop: $A_r \to A_e \to A_c \to A_r \to \dots$ 
At each step $t$, the reasoning state $\bm{s}_t = \{\text{regions}_{1:t-1}, \text{causality}_{1:t-1}, a_{t-1}\}$ updates all previously explored regions, discovered causal relationships and last executed action. 
The model selects the next action $a_t \in \{A_r, A_e, A_c\}$ following the looping order, and then generates an intermediate reasoning result $\bm{r}_t \sim \pi_{\theta}(\cdot \mid \bm{s}_{t}, a_t)$\footnote{$\pi_{\theta}$ refers to the policy of the VLM parameterized by $\theta$.}.
This process iterates until a predefined step limit $T$ is reached. 
The full reasoning trajectory is thus represented as $\bm{\tau} = \bm{r}_1 \oplus \bm{r}_2 \oplus \dots \oplus \bm{r}_T$, which is the accumulation of intermediate results. 
Note that we employ a more capable teacher model to generate high-quality trajectories that better capture causally aware reasoning patterns.

\paragraph{Monte Carlo Tree Search.}
The MCTS algorithm expands a single trajectory into a tree structure, maintaining multiple branches and propagating value estimates to enhance the quality of the optimal trajectory.
Each search iteration consists of four phases: \emph{selection}, \emph{expansion}, \emph{simulation}, and \emph{backpropagation}.
During \emph{selection} phase, the algorithm traverses the tree from the root to identify the most promising child node using the well-known Upper Confidence Bounds for Trees (UCT) rule~\citep{Kocsis2006BanditBM}, which balances exploration and exploitation:
\begin{align}
    \text{UCT}(\bm{s}, a) = Q(\bm{s}, a) + w\sqrt{\frac{\ln N(Pa(\bm{s}))}{N(\bm{s}, a)}},
\end{align}
where $N(\bm{s}, a)$ and $N(Pa(\bm{s}))$ denote the visit counts of the current node and its parent, respectively, and $Q(\bm{s}, a)$ is updated during backpropagation.
In the \emph{expansion} phase, the model executes an action $a$ based on the current node state $\bm{s}$ to generate new child nodes $\mathcal{C}(\bm{s}, a)$, thereby expanding the search tree in both width and depth.
Next, in the \emph{simulation} phase, the model performs an imagined rollout from the selected node until reaching a leaf node. This process estimates a foresighted value by completing a hypothetical reasoning trajectory that anticipates future outcomes.
Finally, during \emph{backpropagation}, the estimated value is propagated upward to update all nodes along the selected branch:
\begin{align}
    Q(\bm{s}, a) = \frac{\sum_{\bm{c} \in \mathcal{C}(\bm{s}, a)} Q(\bm{c}) \cdot N(\bm{c})}{\sum_{\bm{c} \in \mathcal{C}(\bm{s}, a)} N(\bm{c})}.
\end{align}
The reward signal originates from evaluating the leaf node’s state against the ground-truth causal graph, providing supervision for value propagation.
After a fixed number of search iterations, the node values converge, and the optimal trajectory is obtained by greedily selecting the highest-value nodes along the reasoning tree.

\paragraph{Reasoning Policy Initialization.}
\label{subsec:fit}
 
Although the ToCT approach produces controllable reasoning trajectories, it occasionally fails to outperform vanilla one-step reasoning on certain cases.
To ensure that the policy model learns from trajectories that not only reflect structured causally aware reasoning but also lead to superior solutions, we retain only those ToCT-induced trajectories that outperform their vanilla one-step counterparts. 
This guarantees that the model is trained on higher-quality examples, minimizing the risk of introducing suboptimal reasoning patterns. 
We initialize the model's reasoning policy by conducting supervised fine-tuning (SFT) on the filtered trajectories:
\begin{align}
\mathcal{L}_\text{SFT}(\theta) = -\mathbb{E}_{(q, \bm{\tau}) \sim \mathcal{D}_\text{SFT}} \left[ \log \pi_{\theta}(\bm{\tau} \mid q) \right],
\end{align}
Where $q$ denotes the input, and $\bm{\tau}$ is the trajectory to be learned by the policy model $\pi_{\theta}$.

\subsection{Reinforce with Causal Reward}
\label{subsec:rl}

Although imitation learning can instill causally aware reasoning patterns, the VLM’s capability for visual causal discovery remains limited by the narrow coverage of synthetic trajectories.
These trajectories provide useful yet constrained supervision—the model can mimic existing reasoning paths but struggle to generalize.
To address this, we introduce RL to transform the model from a passive imitator into an adaptive causal learner that continually optimizes its causal discovery policy through iterative interaction and feedback.
We start RL from the SFT-trained policy and use the GRPO algorithm.
GRPO dispenses with the separate value function typically required in PPO-style objectives by leveraging relative advantage computed over group sampled outputs for one input.
This design reduces computational overhead and aligns with the visual causal discovery task, where relative comparisons across rollouts capture nuanced improvements in graph reconstruction quality.
Formally, the optimization objective is given by:
\begin{align}
\mathcal{J}_{\text{GRPO}}(\theta) = &\mathbb{E}_{[q \sim \mathcal{D}, \{o_i\} ^G _ {i=1} \sim \pi_{\theta_{\text{old}}} (\cdot|q) ]} \frac{1}{G} \sum ^G _{i=1} ( \notag \\
& \min(R_i A_i, \mathrm{clip}(R_i, 1-\epsilon, 1+\epsilon)A_i)), \notag
\end{align}
\begin{equation}
\resizebox{0.9\linewidth}{!}{$
R_i = \frac{\pi_{\theta}(o_i|q)}{\pi_{\theta_{\text{old}}}(o_i|q)}, \quad A_i=\frac{r_i-\mathrm{mean}(\{r_1,r_2,\dots,r_G\})}{\mathrm{std}(\{r_1,r_2,\dots,r_G\})}.
$}
\end{equation}

\paragraph{Causal Reward Design.}

As outlined in \autoref{sec:task_define}, our goal is to improve the VLM’s ability to understand and reconstruct causal graph structures. Accordingly, we design a causal reward to promote (i) accurate visual causal discovery and (ii) faithful structural generation:
\begin{equation}
\resizebox{0.9\linewidth}{!}{$
R(\bm{\tau})
  = \underbrace{\lambda_r \cdot \text{Recall}(\bm{\tau})
     + \lambda_p \cdot \text{Precision}(\bm{\tau})}_{\text{visual causal discovery}}
     + \lambda_f \cdot \text{Format}(\bm{\tau}).
$}
\end{equation}
The $\text{Recall}$ and $\text{Precision}$ are defined as:
\begin{align}
    \text{Recall} = \frac{|\mathcal{E}_{pred} \cap \mathcal{E}_{gt}|}{|\mathcal{E}_{gt}|}, \text{Precision} = \frac{|\mathcal{E}_{pred} \cap \mathcal{E}_{gt}|}{|\mathcal{E}_{pred}|},
\end{align}
where $\mathcal{E}_{pred}$ and $\mathcal{E}_{gt}$ denote the sets of predicted and ground-truth directed edges, respectively.
By incorporating the $\text{Recall}$ term, the policy model is encouraged to align its predicted causal structure with the ground-truth causal graph.
The $\text{Precision}$ term is introduced to suppress false positives and balance the trade-off between completeness and correctness.
Finally, $\text{Format}$ evaluates the compliance of the model's structural generation with the predefined output schema, which is illustrated in the Appendix.


\begin{table*}[ht]
\centering
\small
\begin{tabular}{lccccccccc}
\toprule
\multirow{2}{*}{\textbf{Model}} &
\multicolumn{3}{c}{\textbf{\data-COCO}} &
\multicolumn{3}{c}{\textbf{\data-365}} &
\multicolumn{3}{c}{\textbf{Average}} \\
\cmidrule(lr){2-4} \cmidrule(lr){5-7} \cmidrule(lr){8-10}
& Recall & Precision & F1 Score & Recall & Precision & F1 Score & Recall & Precision & F1 Score \\
\midrule

Gemini 2.5 Pro & 5.7 & 2.5 & 3.1 & 3.0 & 1.3 & 1.7 & 4.4 & 1.9 & 2.4 \\
OpenAI o3 &  8.6 & 7.7 & 7.0 & 6.0 & 6.4 & 5.1 & 7.3 & 7.1 & 6.1 \\
GPT-4.1 &  16.5 & 15.6 & 15.0 & 3.9 & 3.6 & 3.4 & 10.2 & 9.6 & 9.2 \\
GPT-5 & 5.1 & 2.9 & 3.4 & 3.0 & 1.5 & 1.7 & 4.1 & 2.2 & 3.6 \\

\midrule

Qwen2.5-VL-7B &  4.7 & 7.9 & 5.4 & 2.9 & 6.6 & 3.8 & 3.8 & 7.3 & 4.6 \\
Qwen2.5-VL-32B &  14.0 & 14.1 & 12.3 & 7.7 & 10.3 & 7.7 & 10.9 & 12.2 & 10.0 \\
Qwen2.5-VL-72B &  19.0 & 26.7 & 20.1 & 11.8 & 20.5 & 13.4 & 15.4 & 23.6 & 16.8 \\
\midrule
\multicolumn{1}{l}{} & 
\multicolumn{3}{c}{\textit{in-domain}} & 
\multicolumn{3}{c}{\textit{cross-domain}} \\
\cmidrule(lr){2-4} \cmidrule(lr){5-7}

Qwen2.5-VL-7B + SFT &  11.2 & 14.8 & 12.0 & 12.1 & 19.5 & 13.8 & 11.7 & 17.2 & 12.9 \\
\rowcolor{myspike!20}
\textbf{\model (ours)} &  \textbf{34.2} & \textbf{48.7} & \textbf{37.6} & \textbf{28.1} & \textbf{42.3} & \textbf{31.1} & \textbf{31.2} & \textbf{45.5} & \textbf{34.4} \\
\bottomrule
\end{tabular}
\caption{
\textbf{Comparison of \model with proprietary, open-source, and the SFT variant.(\%) }
We report graph-level Recall, Precision, and F1 Score on both the in-domain (MS-COCO) and cross-domain (Objects365) subsets of \data. \model comprehensively outperform diverse baselines.
}
\vspace{-4mm}
\label{tab:main_results}
\end{table*}

  \section{Experiments}
\label{sec:exp}
\subsection{Setup}
\paragraph{Baselines.}
We compare our \model against three categories of baselines:
(1) state-of-the-art proprietary models, including Gemini 2.5 Pro~\citep{Comanici2025Gemini2P}, OpenAI o3~\citep{OpenAIOA}, and GPT series~\citep{Achiam2023GPT4TR, gpt5};
(2) leading open-source models, represented by the Qwen2.5-VL family~\citep{Bai2025Qwen25VLTR} with varying parameter scales;
(3) an SFT variant\footnote{the SFT variant in the baseline is \textbf{not the same model} as the policy model SFTed on ToCT-induced trajectories, so please do not confuse them.}, directly fine-tuned on ground-truth formatted labels (using the same base model as \model).

\paragraph{Training Data.}
We use the MS-COCO portion of the \data dataset as the training corpus, enabling cross-dataset evaluation on the Objects365 subset. 
The entire training set contains 11,078 samples. 
ToCT is performed on 6,000 of these samples, yielding 3,631 trajectories after filtering. 
The remaining 7,447 samples are used for RL.

\paragraph{Benchmarks and Metrics.}
We evaluate models on 350 samples from the MS-COCO subset of \data for the main benchmark, and another 350 from the Objects365 subset to test cross-dataset performance. For a broader assessment of generalization, we include three out-of-domain (OOD) benchmarks: Math-V and MathVista for multimodal mathematical reasoning, and BLINK for visual perception. We report graph recall, precision, and F1 score computed at a GIoU threshold of 0.5. 

\paragraph{Implementation Details.}
For ToCT, we set the step limit $T$ to 12, expand up to 10 child nodes per step, and run 20 search iterations.
For GRPO, training runs for 15 epochs on 4 nodes with 8 H200 GPUs each, using a batch size of 512 and a maximum response length of 4,096 tokens. Each rollout samples 5 outputs per prompt. 
Further details and prompts templates are provided in the Appendix.

\subsection{Results on \data}
\label{main}
As shown in \autoref{tab:main_results}, \model achieves a substantial improvement over all baselines on both in-domain (\data-COCO) and cross-domain (\data-365) settings.
It surpasses the leading proprietary model (GPT-4.1\footnote{We found that GPT-5 performs poorly in this visually grounded reasoning task, so in subsequent experiments we adopt GPT-4.1.}) by an absolute gain of \textbf{21.0\%} in graph recall, and outperforms the larger open-source model (Qwen2.5-VL-72B) by \textbf{15.8\%}.
\model also maintains strong precision, and a consistent F1 score improvement across both subsets.

Notably, the performance gap is especially evident on the cross-domain \data-365 benchmark.
\model boosts graph recall by \textbf{16.0\%} and F1 score by \textbf{17.3\%} over the SFT variant.
This striking contrast suggests that visually grounded reasoning is the key for VLMs to achieving robust generalization.
Our multi-stage training recipe enables the model to internalize transferrable causal principles, thereby extrapolating causal dependencies to novel scenes, whereas supervised training on totally formatted outputs can severely harm the generalization ability.

The comparison with proprietary models further highlights that, despite their massive scale and pretraining coverage, general-purpose VLMs still struggle to establish coherent cause–effect structures from visual scenes.
In contrast, \model explicitly integrates causal priors and decision optimization through ToCT and RL, enabling a more systematic understanding of visual interactions.
These results validate the effectiveness of our recipe in bridging the gap between perception and reasoning, and underscore that causally aware reasoning is the cornerstone for both interpretability and generalization in visual understanding.

\subsection{Reasoning and Detection}

In this section, we further investigate the respective impact of reasoning and detection capabilities on the overall task performance, primarily reporting the graph recall results on \data-COCO benchmark.

\begin{figure*}[ht]
    \centering   \includegraphics[width=0.95\linewidth]{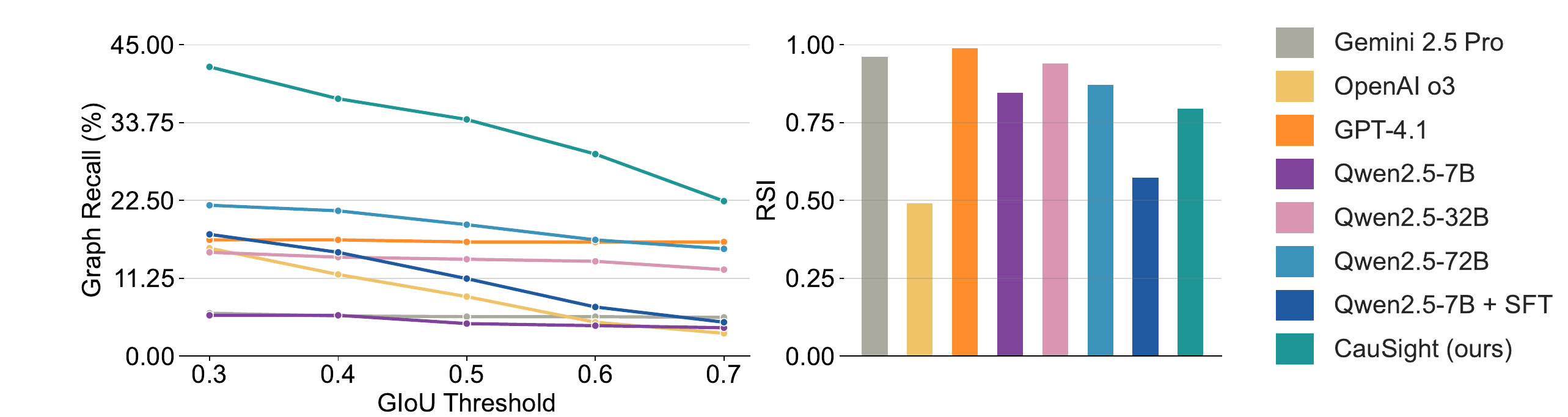}
    \caption{\textbf{Detection-dependent performance stability.} Graph recall is evaluated under varying GIoU thresholds, summarized by the Recall Stability Index (RSI). \model achieves strong causal discovery performance while maintaining high RSI, indicating a balanced integration of detection and reasoning capabilities.
    }
    \vspace{-4mm}
    \label{fig:rsi}
\end{figure*}

\begin{figure}[t]
    \centering
    \includegraphics[width=0.9\linewidth]{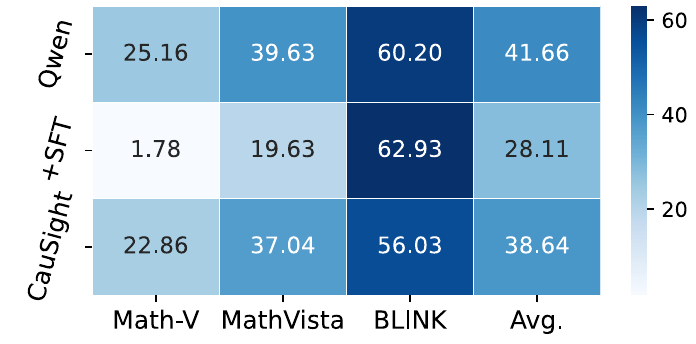}
    \caption{\textbf{Model generalizability across three OOD benchmarks.} Qwen refers to Qwen2.5-VL-7B, +SFT to the SFT variant in the baseline. Each cell corresponds to the model’s accuracy. 
    }
    \vspace{-4mm}
    \label{fig:generalizability}
\end{figure}

\paragraph{Reasoning-Induced Performance Loss.}
\begin{table}[t]
\centering
\small
\begin{tabular}{lccc}
\toprule
\textbf{Model} & \textbf{Recall} & \textbf{$\text{Recall}_{\texttt{R}}$} & \textbf{Loss} \\
\midrule
Gemini 2.5 Pro &  5.7 & 10.33 & 44.8 \\
OpenAI o3 &  8.6 & 9.6 & 10.6 \\
GPT-4.1 &  16.5 & 23.0 & 27.9 \\
Qwen2.5-VL-7B &  4.7 & 5.7 & 18.1 \\
Qwen2.5-VL-32B &  14.0 & 18.2 & 23.1 \\
Qwen2.5-VL-72B &  19.0 & 21.8 & 12.9 \\
\rowcolor{myspike!20}
\textbf{\model (ours)} &  \textbf{34.2} & \textbf{37.2} & \textbf{8.0} \\
\bottomrule
\end{tabular}
\caption{
\textbf{Causal Reasoning-Induced Performance Loss.(\%)} \model effectively mitigates reasoning-induced loss toward reachable upper bound.
}
\vspace{-4mm}
\label{tab:abla_2}
\end{table}
We isolate the impact of reasoning from detection by evaluating it in a controlled setting.
Specifically, for each causal graph, we compute a \emph{reachable} $\text{Recall}_{\mathcal{R}}$, defined as the maximum achievable recall given the set of correctly recognized entities (i.e., assuming perfect causally aware reasoning over detected nodes).
Let $\text{Recall}$ denote the actual graph recall obtained by the model. We then quantify the reasoning-induced performance loss as:
\begin{align}
\mathcal{L} = \frac{\text{Recall}_\mathcal{R} - \text{Recall}}{\text{Recall}_\mathcal{R}}.
\end{align}
This metric reflects the relative gap between the model’s achievable upper bound and its actual causally aware reasoning capability.
As shown in \autoref{tab:abla_2}, the reasoning gap markedly affects the recall performance across models\footnote{Since the model fine-tuned directly on ground-truth labels (the SFT variant) follows a structured output and no longer falls under the category of \textbf{reasoning models}, its relevant results are not reported.}.
While larger models such as Qwen2.5-VL-72B and GPT-4.1 achieve solid graph recall, they still suffer substantial performance loss toward the reachable upper bounds, suggesting insufficient causal comprehension.
In contrast, \model achieves a recall of \textbf{34.2\%}, outperforming even much larger models while reducing the loss to \textbf{8.0\%}, effectively mitigating reasoning-induced degradation.

\paragraph{Detection-Dependent Performance Stability.}

It is inherently challenging to evaluate a model’s detection capability in isolation on this task, since the detected entities are closely intertwined with their causal relations. 
To partially disentangle these factors, we evaluate graph recall under different GIoU thresholds (ranging from 0.3 to 0.7), which reflects its sensitivity to detection strictness, thus indicating models' underlying detection capability. 
Meanwhile, we introduce \emph{Recall Stability Index} $\text{(RSI)}$, denoted as: 
\begin{align}
\text{RSI}_{[0,1]} = \max\left(0, 1 - \frac{\mathrm{std}(\{\text{Recall}_t\})}{\mathrm{mean}(\{\text{Recall}_t\})}\right),
\end{align}
where $\text{Recall}_t$ denotes graph recall at GIou threshold $t$.
RSI quantifies how robust graph recall is across thresholds, i.e., how strongly it depends on detection capability.
As \autoref{fig:rsi} shows, GPT-4.1 exhibits highest stability, indicating superior detection capability together with competitive recall. 
In contrast, directly applying SFT on the base model severely degrades detection capability, as reflected by the markedly low RSI.
\model achieves a balanced improvement: it substantially boost causal discovery performance with only a minor drop in RSI.

To summarize, by comparing the reasoning-induced loss with RSI, we observe an opposite trend between reasoning and detection capabilities. 
For example, OpenAI o3 shows a low reasoning-induced performance loss (\textbf{10.6\%}), yet exhibits the weakest RSI; conversely, Gemini 2.5 Pro demonstrates excellent detection performance but suffers an extremely high reasoning-induced loss (\textbf{44.8\%}), suggesting an absence of causally aware reasoning ability. 
These findings highlight that connecting visual perception to reasoning remains a challenging problem. 
Meanwhile, \model not only achieves leading visual causal discovery performance but also maintains a harmonious balance between detection and reasoning.

\begin{figure*}[t]
    \centering   \includegraphics[width=0.95\linewidth]{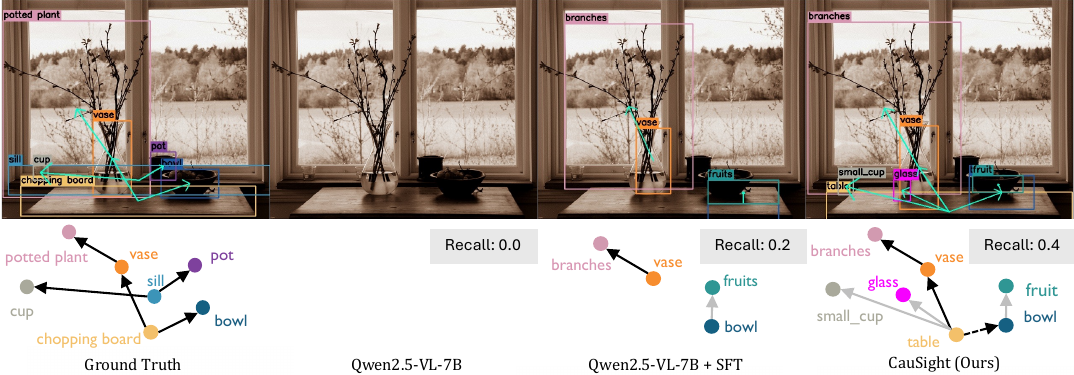}
    \caption{\textbf{Qualitative comparison on the base model, the SFT variant and our \model. }}
    \vspace{0mm}
    \label{fig:quality_compare}
\end{figure*}

\begin{table*}[ht]
\centering
\small
\resizebox{0.95\textwidth}{!}{
\begin{tabular}{lccccccccc}
\toprule
\multirow{2}{*}{\textbf{Model}} &
\multicolumn{3}{c}{\textbf{\data-COCO}} &
\multicolumn{3}{c}{\textbf{\data-365}} &
\multicolumn{3}{c}{\textbf{Average}} \\
\cmidrule(lr){2-4} \cmidrule(lr){5-7} \cmidrule(lr){8-10}
 & Recall & Precision & F1 Score & Recall & Precision & F1 Score & Recall & Precision & F1 Score \\
\midrule
\rowcolor{myspike!20}
\textbf{\model (ours)} &  \textbf{34.2} & 48.7 & \textbf{37.6} & \textbf{28.1} & \textbf{42.3} & \textbf{31.1} & \textbf{31.2 (+27.4)} & \textbf{45.5 (+38.2)} & \textbf{34.4 (+29.8)} \\
w/o ToCT &  31.0 & \textbf{50.9} & 35.3 & 18.0 & 36.5 & 21.9 & 24.5 (+20.7) & 43.7 (+36.4) & 28.6 (+24.0) \\
w/o RL &  10.7 & 6.8 & 7.3 & 9.4 & 6.3 & 6.6 & 10.1 (+6.3) & 6.6 (-0.7) & 7.0 (+2.4) \\
\midrule
Qwen2.5-VL-7B &  4.7 & 7.9 & 5.4 & 2.9 & 6.6 & 3.8 & 3.8 & 7.3 & 4.6 \\
\bottomrule
\end{tabular}
}
\caption{\textbf{Ablation results on the training recipe.(\%)} Both the ToCT approach and the subsequent RL stage play critical roles in enhancing the performance of \model. Integrating SFT on ToCT-induced trajectories before RL yields a evident advantage over applying RL alone, especially on the cross-domain benchmark \data-365.}
\vspace{-2mm}
\label{tab:abla_1}
\end{table*}

\subsection{Ablation Study on Training Recipe}

We conduct an ablation study on our training recipe, evaluating on both the in-domain (\data-COCO) and cross-domain (\data-365) benchmarks.
As shown in \autoref{tab:abla_1}, both the ToCT approach and the subsequent RL stage contribute substantially to the overall performance gains.
The introduction of ToCT-induced trajectories provide a supervised prior that enables the model to initialize a structured reasoning policy, while RL further optimizes the policy through dynamic feedback.
On average, \model improves graph recall by \textbf{27.4\%} over the base model, and by \textbf{21.1\%} and \textbf{6.7\%} over the model without RL and that without ToCT, respectively.
%
Notably, despite the limited number of ToCT-induced trajectories, initializing the model with SFT on them before RL resulted in a marked advantage over using RL alone, with a \textbf{10.1\%} increase in graph recall on the cross-domain benchmark.
This again underscores that enhanced reasoning is key to achieving better generalization as discussed in \autoref{main}.
Quality analysis of the synthesized trajectories are also provided in the Appendix.

\subsection{General Capability Assessment}

To evaluate the generalizability of \model, we compare it to the base model and the SFT variant on three benchmarks: Math-V and MathVista for multimodal mathematical reasoning, and BLINK for visual perception. \autoref{fig:generalizability} presents the results. Our findings indicate that \model largely preserves generalization capabilities comparable to the original model. Conversely, models trained directly with SFT exhibit a severe degradation. This is particularly evident on math reasoning tasks, where the performance of the SFT variant dropped by over \textbf{20\%}.

\subsection{Qualitative Comparison}
\autoref{fig:quality_compare} presents a qualitative comparison among the base model, the SFT variant, and our \model. \model achieves high-fidelity causal relation detection, producing detailed and structurally consistent outputs. Additionally, under our multi-stage training using graph reward without category-label supervision, the model is able to perform genuine attribute-based reasoning. For example, it can correctly revise the ground-truth label \texttt{chopping board} to \texttt{table} for causal discovery. These results highlight its effectiveness in capturing complex causal patterns within visual scenes.

\section{Related Work}

\paragraph{Causal Discovery and Reasoning.}
Causal reasoning has been widely explored with structural causal models (SCMs)~\citep{pearl2009causality} offering a formal basis.
Classical approaches~\citep{Shimizu2006ALN, Peters2014UvADARED} work well in low-dimensional settings but scale poorly to complex data.
Recent studies combine deep learning with causal modeling for induction from interactive data~\citep{Dasgupta2019CausalRF, Buesing2018WouldaCS, Nair2019CausalIF}, though mostly in synthetic domains.
Graph-based and differentiable DAG methods~\citep{Yu2019DAGGNNDS, Lachapelle2019GradientBasedND} improve scalability but rarely model explicit causal structures from visual observations.
Our work directly learns interpretable causal graphs from raw images in the vision-language context.

\paragraph{Causal Evaluation in LLMs and VLMs.}
LLMs have been tested on causal reasoning in text—covering commonsense, extraction, and multi-step reasoning~\citep{DBLP:conf/aaai/SapBABLRRSC19, Chen2022ERGOER, Chen2023LearningTT, Lyu2020ReasoningAG}—but visual grounding remains limited.
Existing visual QA and causal benchmarks~\citep{Zhu2015Visual7WGQ, Goyal2016MakingTV, Wang2016FVQAFV, Zellers2018FromRT, Park2020VisualCOMETRA} touch on causality but lack explicit structures and rely on shallow event reasoning.
We address this gap by enabling explicit causal graph learning from images for fine-grained, entity-level reasoning in vision-language contexts.
  \section{Conclusions}

This work introduces \model, a novel VLM capable of constructing causal graphs across diverse visual scenes. 
We formulate the task of \emph{visual causal discovery} and propose the first large-scale Visual Causal Graph dataset (\data), to support further research in this field. 
We believe that causal understanding is pivotal for AI systems to truly generalize to complex environments and unseen tasks.

  \section*{Acknowledgement}
\label{sec:ack}
This work is supported by Shanghai Artificial Intelligence Laboratory.

  \bibliography{src/custom}
  \bibliographystyle{ieeenat_fullname}
\fi

\appendix
\label{sec:appendix}

\maketitlesupplementary

\section{Synthesized Trajectories}

\begin{figure*}[t]
    \centering
    \includegraphics[width=1\linewidth]{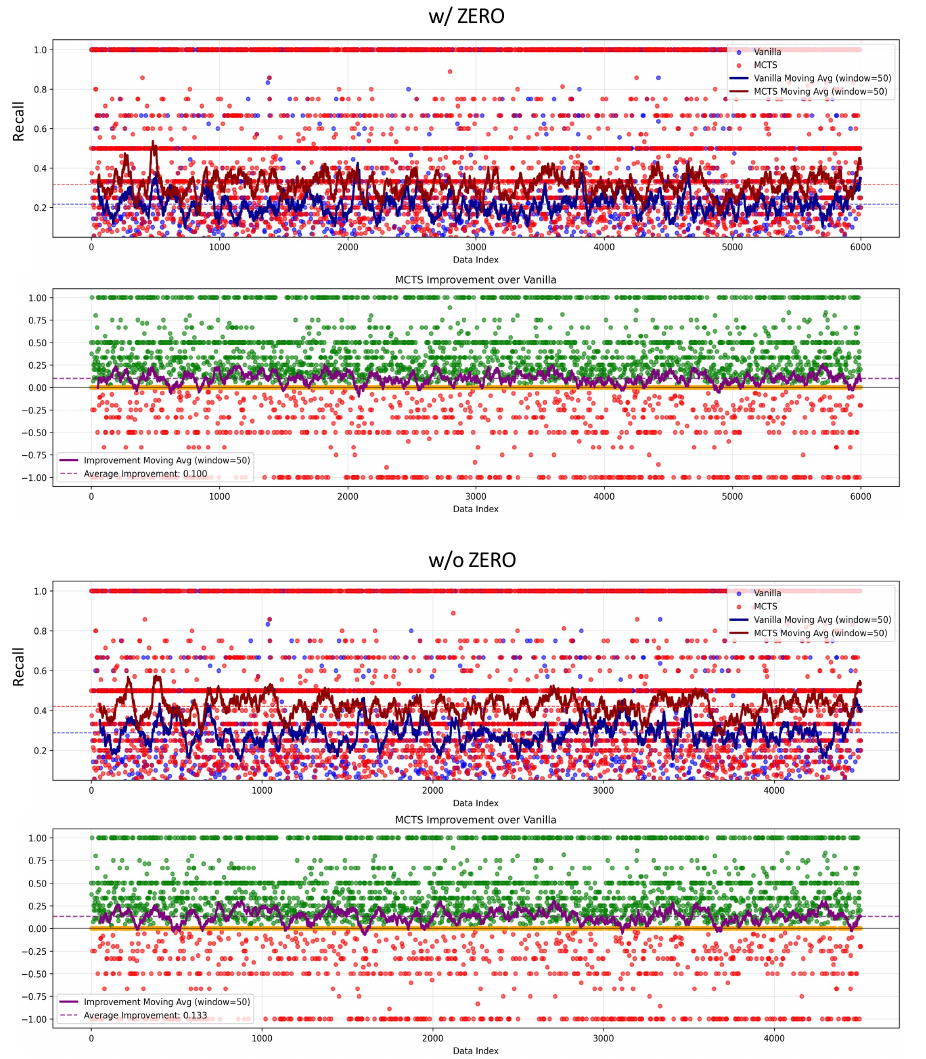}
    \caption{\textbf{Specific results of 6000 samples (ToCT vs. vanilla, scatter + moving average, w/ ZERO and w/o ZERO.)} ZERO represents samples where both the ToCT and the vanilla approaches produce solutions with 0 recall.}
    \label{fig:scatter}
\end{figure*}

In this section, we present a detailed analysis on the quality of ToCT-induced reasoning trajectories.
As discussed in \autoref{subsec:fit}, the ToCT approach could produce suboptimal solutions compared to their vanilla one-step reasoning counterparts on certain cases.
We provide the overall quantification results in \autoref{tab:quality}.
The \textbf{ZERO} represents samples where both the ToCT and the vanilla approaches produce solutions with \textbf{0 recall}.
These samples are regarded as relatively \emph{extreme cases}, e.g., those with excessively high difficulty.
We can see that after removing the extreme cases (w/o ZERO), mean recall of ToCT achieves 0.42, with a 13\% absolute gain over the vanilla approach.
We further analyzed the data and found that about 50\% of the samples had an $> 0.1$ improvement (ToCT over vanilla).
\autoref{fig:scatter} shows the specific results of 6,000 samples (scatter).

Our rule for filtering high-quality trajectories is that the ToCT approach is required to strictly outperform the vanilla one-step reasoning on the same image. 
This ensures that filtered trajectories not only follow the designed reasoning patterns but also lead to superior solutions, which minimize the risk of introducing suboptimal reasoning paths.
Finally, after the filtering, we retain 3,631 from the total 6,000 trajectories.
The mean recall of the filtered trajectories is 0.50, while that of their vanilla counterparts is 0.21. 
\begin{table}[ht]
\centering
\begin{tabular}{lcccc}
\toprule
 & \multicolumn{2}{c}{\textbf{w/ ZERO}} & \multicolumn{2}{c}{\textbf{w/o ZERO}} \\
\cmidrule(lr){2-3} \cmidrule(lr){4-5}
\textbf{Metric} & \textbf{Vanilla} & \cellcolor{myspike!20}\textbf{ToCT} & \textbf{Vanilla} & \cellcolor{myspike!20}\textbf{ToCT} \\
\midrule
Mean & 0.22 & \cellcolor{myspike!20}\textbf{0.32} & 0.29 & \cellcolor{myspike!20}\textbf{0.42} \\
Median & 0.00 & \cellcolor{myspike!20}\textbf{0.22} & 0.13 & \cellcolor{myspike!20}\textbf{0.33} \\
\bottomrule
\end{tabular}


\caption{\textbf{Recall comparison between ToCT and vanilla approaches under different conditions.} The ToCT approach is overall superior to vanilla one-step reasoning.}
\label{tab:quality}
\end{table}

\begin{figure*}[ht]
    \centering
    \includegraphics[width=1\linewidth]{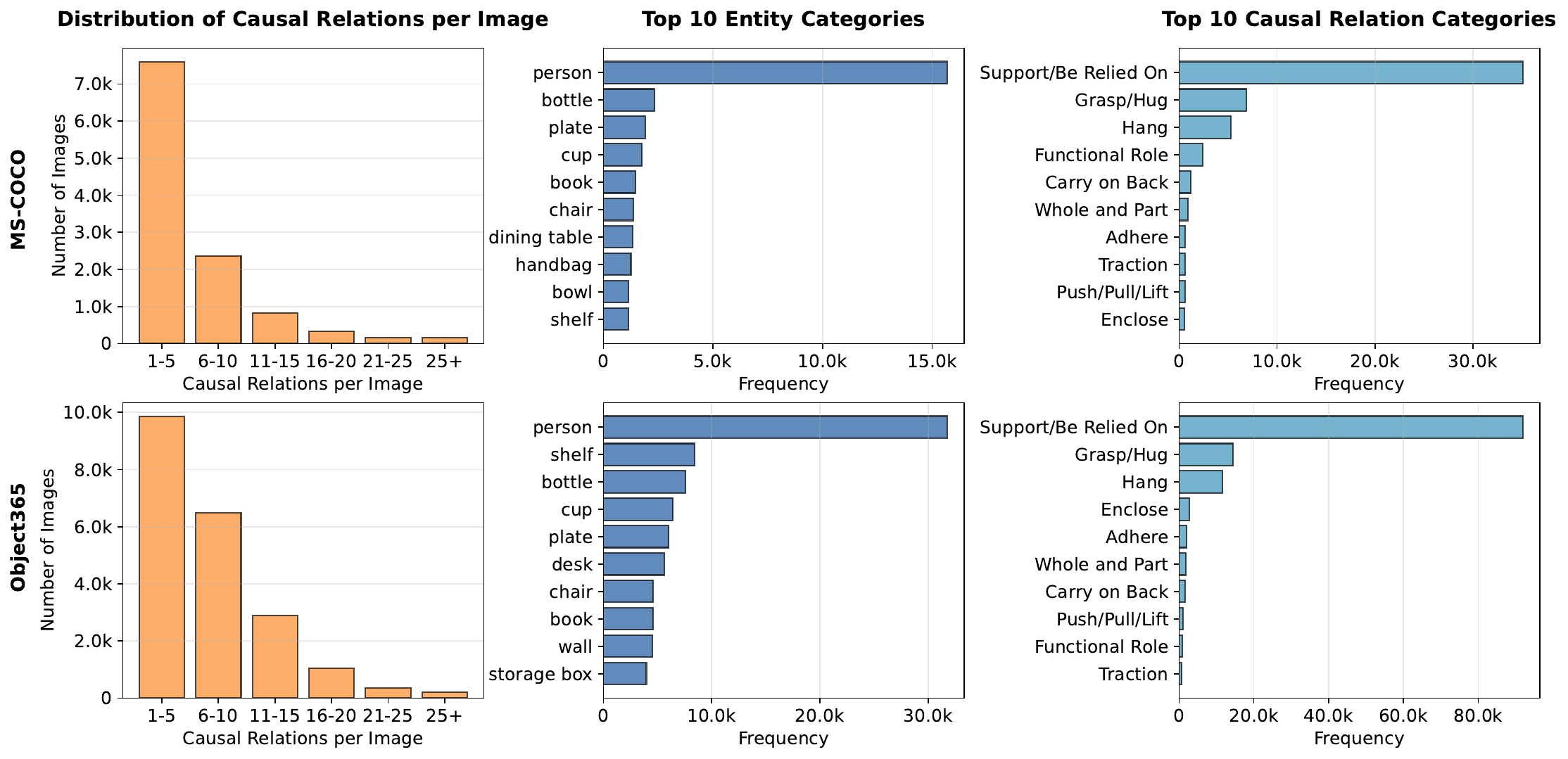}
    \caption{\textbf{Dataset statistics.}}
    \label{fig:data_stats}
\end{figure*}

\section{Dataset}
We sample candidate images from the MS-COCO and Object 365 datasets based on the number of entities present in each image and the diversity of their associated labels. The annotation process consists of two stages: bounding box refinement and causal-relationship labeling.

\paragraph{Bounding box refinement requirements.} To define the target objects for annotation, all entities present in the image are considered, including both objects and humans. Each entity must be annotated individually and cannot be merged with others. Entities smaller than 30×30 pixels are excluded.
The annotation process is based on existing pre-annotated bounding boxes. Annotators are required to review these boxes and modify, remove, or add annotations according to the current guidelines. Pre-annotated bounding boxes should be corrected or deleted under three conditions. (1) Boundary errors: the bounding box contains excessive empty space or truncates part of the entity. (2) Label errors: the assigned label is overly broad or abstract (e.g., “sky,” “air”) or is factually incorrect (e.g., labeling a man as a woman). (3) Irrelevant entities: entities that have no direct physical contact or interaction with any other entity in the scene.
Besides, new entities should be added only if they meet the following criterion: the entity engages in direct physical contact or force interaction with at least one other entity in the scene, but is not annotated by the existing dataset.

\paragraph{Causal relationship labeling requirements.} Given images with annotated bounding boxes and entity labels, annotators identify causal relationship attributes and directions for every ordered pair of entities and assign each relationship to one of the predefined categories, such as ``carry on'' and ``support''.
A causal relationship $i\to j$ exists when three conditions are simultaneously met: (1) $\mathbf{v}_i$ is in direct contact with $\mathbf{v}_j$, (2) the presence of $\mathbf{v}_i$ maintains the current state of $\mathbf{v}_j$, (3) removing $\mathbf{v}_i$ would cause $\mathbf{v}_j$ to lose its current state. 

To ensure the quality, we start the annotation process after all the annotators have completed the training and achieved an accuracy rate of 95\% in the pre-annotated validation set. In total, 50 junior annotators participated in the labeling process, and 10 senior annotators conducted quality reviews. The senior annotators examined whether the annotations produced by junior annotators in both stages adhered to the corresponding guidelines, thereby further guaranteeing the overall quality of our VCG-32K dataset.
Figure \ref{fig:data_stats} shows the distribution of causal relations per image, the top 10 entity categories, and the top 10 causal relation categories for the MS-COCO and Objects365 subset of the final VCG-32K dataset, respectively. 
An example of VCG-32K is shown below, where a causal relationship is denoted by entity IDs, with the first entity causing the second entity.
{\small\begin{verbatim}
{
    "dataset_id": "COCO", 
    "img_id": 0, 
    "entities": [    
        {
            "entity_id": 0,  
            "entity_name": "woman", 
            "bbox": [502.6, 105.47, 
            25.83, 132.38]  
        },
        ...
    ],
    "causal_relationships":[  
        "carry_on": [[0, 1], ...],  
        "support": [[2, 3], ...]
        ...
    ]
}
\end{verbatim}}


\section{Qualitative Comparisons}
We provide more qualitative comparisons on Qwen2.5-VL-7B (base model), Qwen2.5-VL-7B + SFT and our \model.
For each, we present the ultimate generated causal graph and corresponding recall against the ground-truth.
Black solid lines denote correctly constructed causal edges, black dashed lines indicate correctly inferred causal relationships but without precise bounding box matches, and gray solid lines represent incorrectly constructed causal edges.

\begin{figure*}[ht]
    \centering
    \includegraphics[width=1\linewidth]{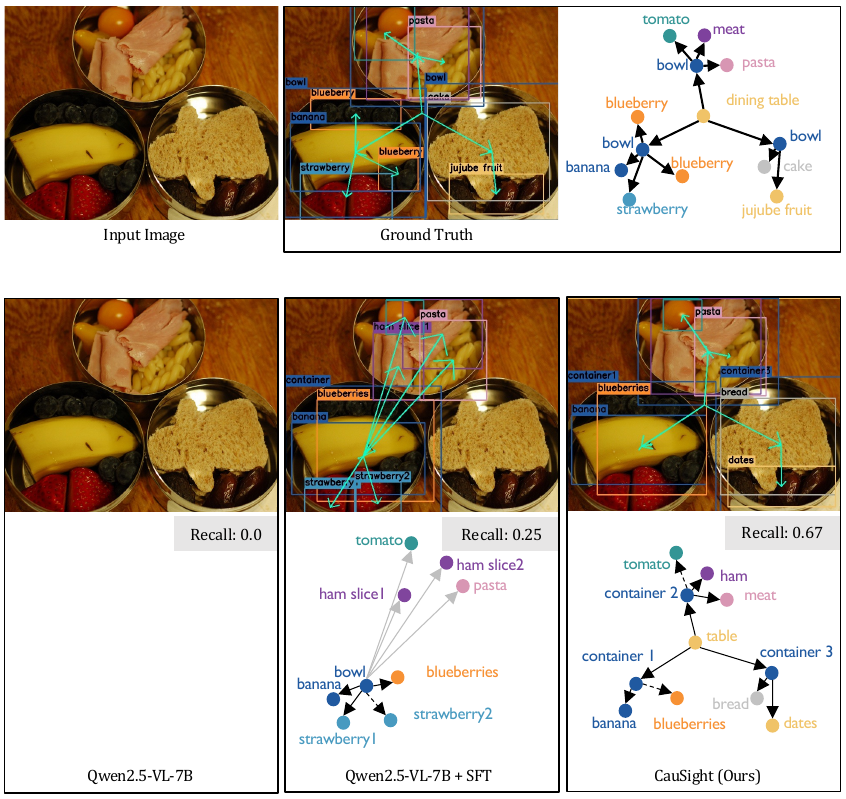}
    \caption{\textbf{Qualitative comparison (1).}}
    \label{fig:case1}
\end{figure*}

\begin{figure*}[ht]
    \centering
    \includegraphics[width=1\linewidth]{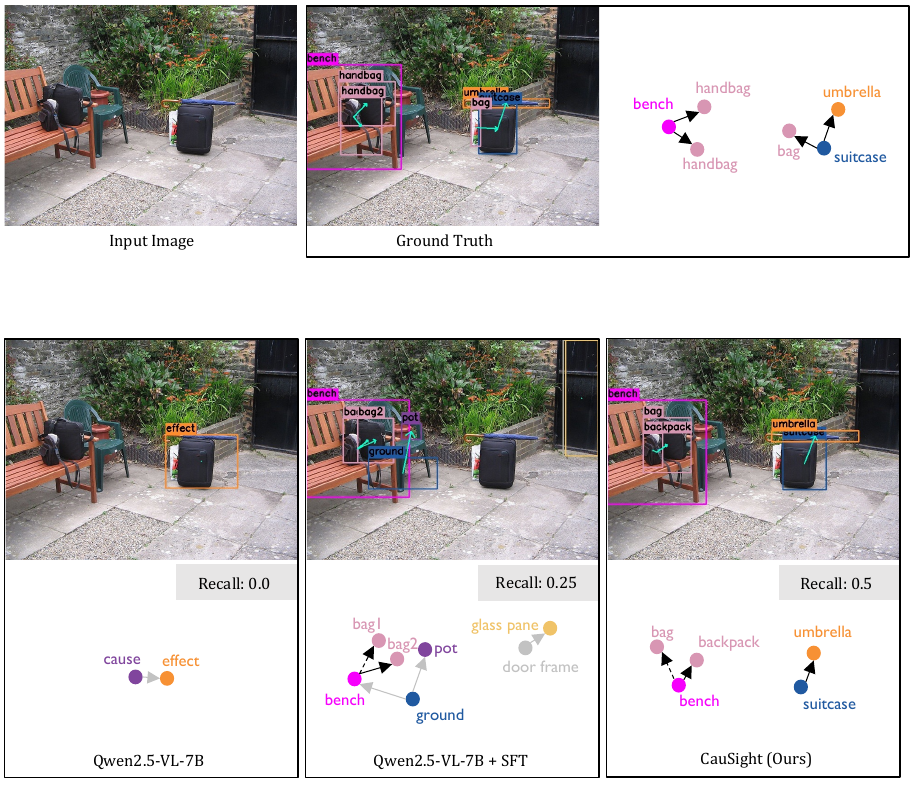}
    \caption{\textbf{Qualitative comparison (5).}}
    \label{fig:case5}
\end{figure*}

We observe that the base model struggles to construct a causal graph in most cases, due to its limited detection and causal discovery capabilities. In our provided cases, the base model consistently achieves a final recall of 0.
The model SFTed with ground-truth labels shows slight improvement; however, it exhibits severe hallucinations in complex scenarios as \autoref{fig:case1} and \autoref{fig:case5} show (cases 1 and 5). For example, it incorrectly assumes that one container causally affects the food inside another container. In simpler scenarios, it tends to over-reason: although it achieves a recall of 0.5 in cases 3 and 4, its precision remains extremely low because it falsely detects many nonexistent entities and infers their nonexistent causal relationships. 
Our model, \model, after multi-stage training, demonstrates strong performance. It reliably captures the essential causal structure of a scene. In more complex cases (1, 2, and 5), although the recall does not reach 1, it correctly identifies the main causal relationships (some correct edges are counted as false due to imperfect bounding box matching, as indicated by the black dashed lines). Meanwhile, in the simpler scenarios (3 and 4), the model even achieves perfect recall, showing strong generalization even under a complex reasoning pattern.

\begin{figure*}[ht]
    \centering
    \includegraphics[width=0.9\linewidth]{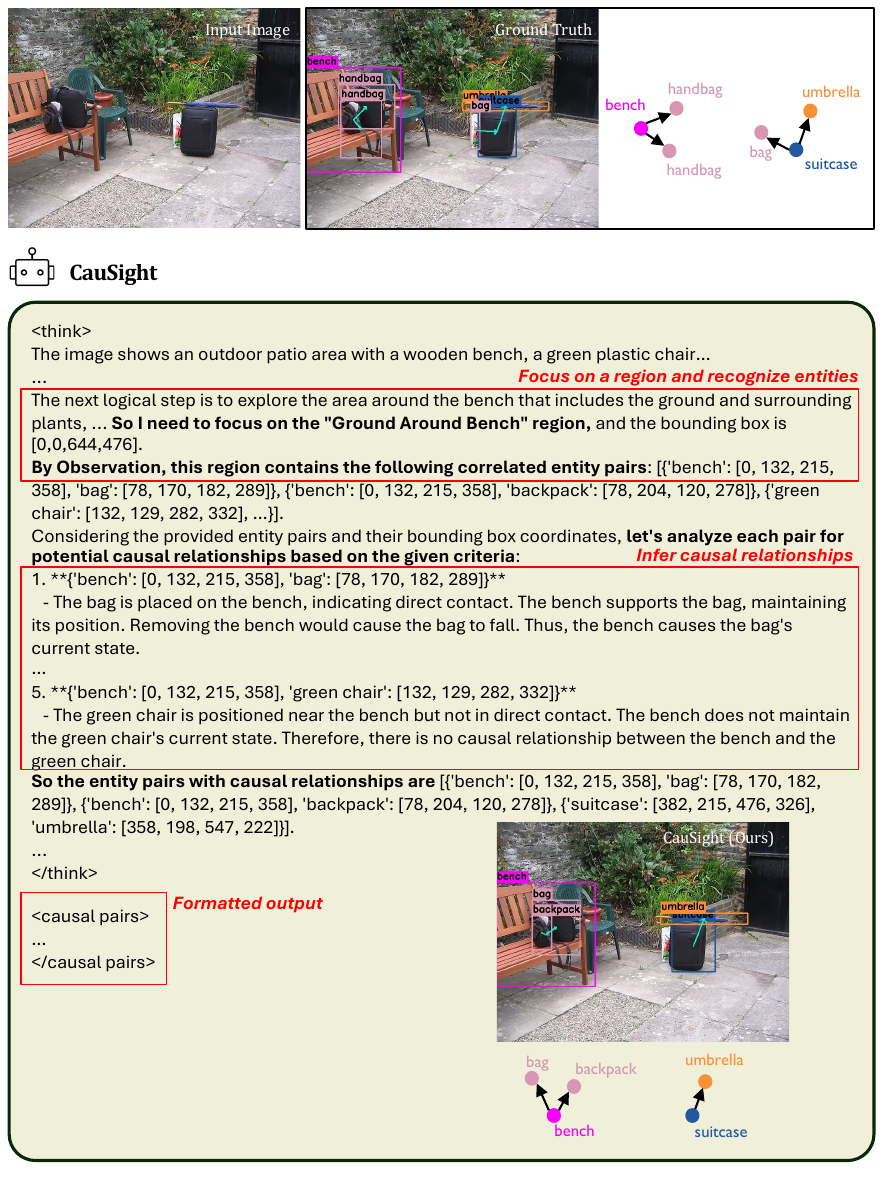}
    \caption{\textbf{Illustration of actual reasoning path from \model.}}
    \label{fig:case}
\end{figure*}

Additionally, we provide an actual reasoning path from \model as shown in \autoref{fig:case}. 
The reasoning path is still well structured to include steps like ``focus on a region'', ``recognize entities'' and ``infer causal relationships'', following the causally aware reasoning patterns we designed in \autoref{subsec:toct}.  

\clearpage

\begin{figure*}[ht]
    \centering
    \includegraphics[width=1\linewidth]{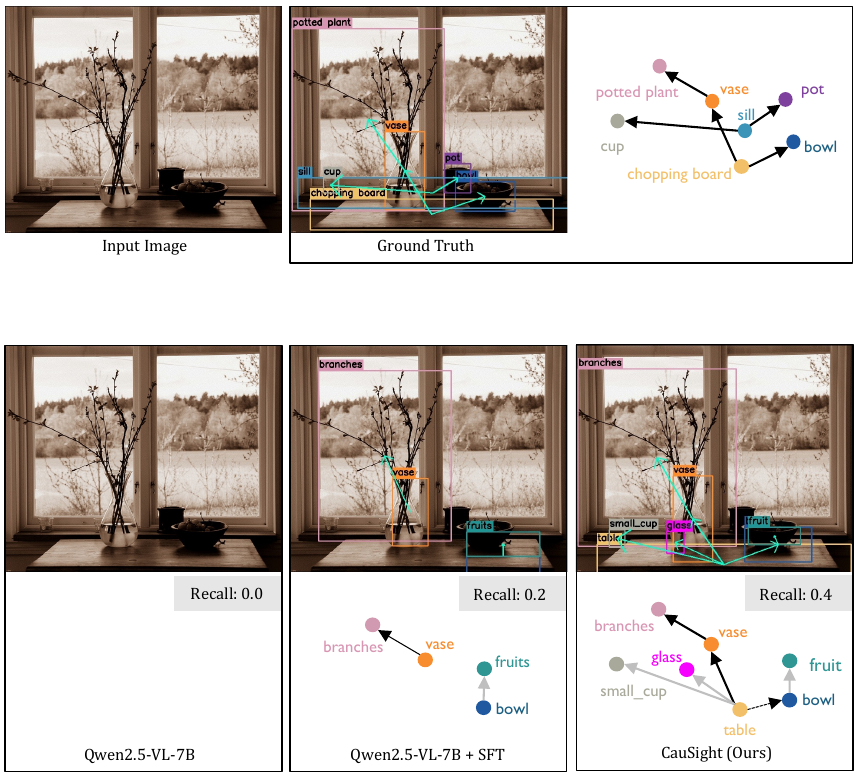}
    \caption{\textbf{Qualitative comparison (2).}}
    \label{fig:case2}
\end{figure*}

\begin{figure*}[ht]
    \centering
    \includegraphics[width=1\linewidth]{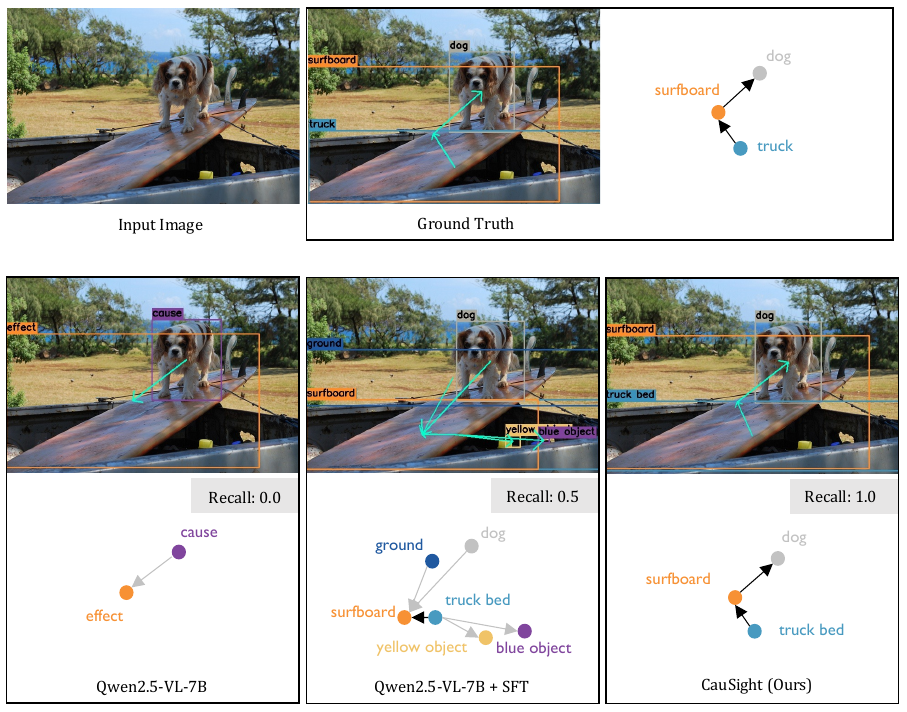}
    \caption{\textbf{Qualitative comparison (3).}}
    \label{fig:case3}
\end{figure*}

\begin{figure*}[ht]
    \centering
    \includegraphics[width=1\linewidth]{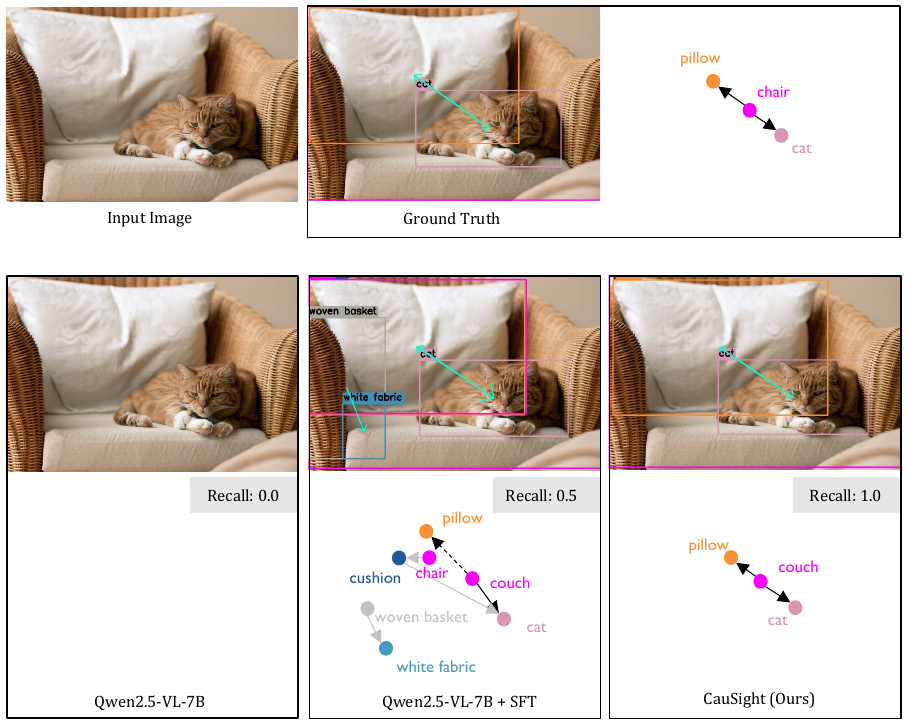}
    \caption{\textbf{Qualitative comparison (4).}}
    \label{fig:case4}
\end{figure*}

\clearpage

\clearpage
\onecolumn
\begin{tcolorbox}[
  width=\textwidth,
  colback=gray!5,
  colframe=black,
  title=\texttt{Region Selection},
  coltitle=black,
  colbacktitle=mygreen!20,
  fonttitle=\large\bfseries,
  boxrule=0.8pt,
  arc=4pt,
  left=6pt,
  right=6pt,
  top=6pt,
  bottom=6pt,
  fontupper=\ttfamily, 
]
You are analyzing the causal relationships between entities in the image through multiple steps.\\
Your current reasoning trajectory is as follows:

Explored regions: \textcolor{myorange}{\{explored regions\}}.

Identified causal pairs: \textcolor{myorange}{\{causal pairs\}}.

Now we hope to look for new regions to discover more potential correlated entity pairs.\\
Please select the next most worthy region to focus on and explain your thinking process.\\
Note: the next region should be DIFFERENT from the previous explored regions.

\medskip
\noindent -- If you think the exploration regions and identified causal pairs are SUFFICIENTLY COMPREHENSIVE, you should DIRECTLY output ``END TRACE'' and nothing else.

\medskip
\noindent Otherwise, your output format should be as follows:

\medskip
\texttt{<think>}\\
(State the reason as concisely as possible for selecting the new focused region.)\\
\texttt{</think>}

\medskip
\texttt{<region name>}\\
(Output the name of the focused region and nothing else.)\\
\texttt{</region name>}

\medskip
\texttt{<bounding box>}\\
(Output the bounding box of the focused region with format [x1, y1, x2, y2] and nothing else, where (x1, y1) is the top-left coordinate and (x2, y2) is the bottom-right coordinate of the bounding box.)\\
\texttt{</bounding box>}
\end{tcolorbox}
\vspace{20pt}
\begin{tcolorbox}[
  width=\textwidth,
  colback=gray!5,
  colframe=black,
  title=\texttt{Entity Recognition},
  coltitle=black,
  colbacktitle=mygreen!20,
  fonttitle=\large\bfseries,
  boxrule=0.8pt,
  arc=4pt,
  left=6pt,
  right=6pt,
  top=6pt,
  bottom=6pt,
  fontupper=\ttfamily, 
]
Your task is to identify all entity pairs that may have correlations in the image.\\
Each pair should have obvious potential correlations such as spatial dependence, support, grasping, placement, inclusion, etc.\\
Think and output all these correlated entity pairs and their bounding boxes.

\medskip
\noindent Your output format should be as follows:

\medskip
\texttt{<think>}\\
(Provide the concise thinking process for identifying correlated entity pairs.)\\
\texttt{</think>}

\medskip
\texttt{<entity pairs>}\\
(Output all the correlated entity pairs in the format of ``[{"entity1": [x1, y1, x2, y2], "entity2": [x1, y1, x2, y2]}, {"entity3": [x1, y1, x2, y2], "entity4": [x1, y1, x2, y2]}, ...]''. You should use ACTUAL ENTITY NAME to replace the placeholders ``entity1'', ``entity2'', ... in the format. (x1, y1) is the top-left coordinate and (x2, y2) is the bottom-right coordinate of the bounding box.)\\
\texttt{</entity pairs>}

\end{tcolorbox}
\clearpage
\begin{tcolorbox}[
  width=\textwidth,
  colback=gray!5,
  colframe=black,
  title=\texttt{Causality Orientation},
  coltitle=black,
  colbacktitle=mygreen!20,
  fonttitle=\large\bfseries,
  boxrule=0.8pt,
  arc=4pt,
  left=6pt,
  right=6pt,
  top=6pt,
  bottom=6pt,
  fontupper=\ttfamily, 
]
Based on the image, your task is to determine whether causal relationships exist between the following entity pairs.\\
Entity pairs: \textcolor{myorange}{\{entity pairs\}}

\medskip
The causality criteria are as follows:\\
For example, if the entity pairs are \{\{"A": [x1, y1, x2, y2], "B": [x1, y1, x2, y2]\}\} or \{\{"B": [x1, y1, x2, y2], "A": [x1, y1, x2, y2]\}\}:\\
- A is in direct contact with B. \\
- A's presence maintains B's current state. \\
- Removing A would cause B to lose its current state.\\
Then A is the cause and B is the effect.\\
(x1, y1) is the top-left coordinate and (x2, y2) is the bottom-right coordinate of the bounding box.

\medskip
Your output format should be as follows:

\medskip
\texttt{<think>}\\
(Consider entity pairs and keep the reasoning as concise as possible.)\\
\texttt{</think>}

\medskip
\texttt{<causal pairs>}\\
(Output entity pairs with causal relationships only and if necessary, swap the ORDER of entities pairs to ensure the cause precedes the effect.)\\
\texttt{</causal pairs>}

\end{tcolorbox}
\vspace{20pt}
\begin{tcolorbox}[
  width=\textwidth,
  colback=gray!5,
  colframe=black,
  title=\texttt{\textbf{E2E} Visual Causal Discovery},
  coltitle=black,
  colbacktitle=mygreen!20,
  fonttitle=\large\bfseries,
  boxrule=0.8pt,
  arc=4pt,
  left=6pt,
  right=6pt,
  top=6pt,
  bottom=6pt,
  fontupper=\ttfamily, 
]

Identify all causal relationships between entities in the image based on the following criteria:\\
- A is in direct contact with B.\\
- A's presence maintains B's current state.\\
- Removing A would cause B to lose its current state.\\
Then A is the cause and B is the effect.

\medskip
\noindent Please provide your reasoning process and output all the entity pairs with causal relationships and their bounding boxes in the following format:

\medskip
\texttt{<think>}\\
(Provide your reasoning process for analyzing the image.)\\
\texttt{</think>}

\medskip
\texttt{<causal pairs>}\\
(Output all the entity pairs with causal relationships and their bounding boxes in the format of ``[{"cause": [x1, y1, x2, y2], "effect": [x1, y1, x2, y2]}, {"cause": [x1, y1, x2, y2], "effect": [x1, y1, x2, y2]}, ...]''. You should use ACTUAL ENTITY NAME to replace the placeholders ``cause'' and ``effect'' in the format. (x1, y1) is the top-left coordinate and (x2, y2) is the bottom-right coordinate of the bounding box.)\\
\texttt{</causal pairs>}
\end{tcolorbox}

\end{document}